%% file: main_final.tex
\documentclass[10pt,twocolumn,letterpaper]{article}

\usepackage[sort&compress,square,comma,numbers]{natbib}

\usepackage{cvpr}
\usepackage{times}
\usepackage{epsfig}
\usepackage{graphicx}
\usepackage{amsfonts}
\usepackage{amsmath}
\usepackage{amssymb}
\usepackage{amsthm}
\usepackage{mathtools}

\usepackage{nicefrac}       %
\usepackage{microtype}      %
\usepackage{booktabs}  
\usepackage{multirow}
\usepackage{multicol}
\usepackage[capitalise]{cleveref}
\usepackage{xcolor}
\usepackage{bm}
\usepackage[british]{babel}

\newcommand{\xpm}[1]{{\tiny$\pm#1$}}

\newcommand\scalemath[2]{\scalebox{#1}{\mbox{\ensuremath{\displaystyle #2}}}}

\usepackage{caption}

\newcommand\blfootnote[1]{%
  \begingroup
  \renewcommand\thefootnote{}\footnote{#1}%
  \addtocounter{footnote}{-1}%
  \endgroup
}

\usepackage[pagebackref=true,breaklinks=true,letterpaper=true,colorlinks,bookmarks=false]{hyperref}

\cvprfinalcopy %

\usepackage{cuted}
\usepackage{capt-of}

\begin{document}

\title{Semantic Image Manipulation Using Scene Graphs}

\author{Helisa Dhamo $^{1,}$ \thanks{The first two authors contributed equally to this work} \\
\and
Azade Farshad $^{1,}$ \footnotemark[1]\\
\and
Iro Laina $^{1,2}$\\
\and
\and
Nassir Navab $^{1,3}$\\
\and
Gregory D. Hager $^{3}$\\
\and
Federico Tombari $^{1,4}$\\
\and
Christian Rupprecht $^{2 }$\\ 
\and 
$^{1}$ Technische Universit\"at M\"unchen
\hspace{0.25cm}
$^{2}$ University of Oxford
\hspace{0.25cm}
$^{3}$ Johns Hopkins University
\hspace{0.25cm}
$^{4}$ Google
}

\maketitle
\begin{strip}
\begin{center}
\centering
\vspace{-1.5cm}
\includegraphics[width=0.9\textwidth]{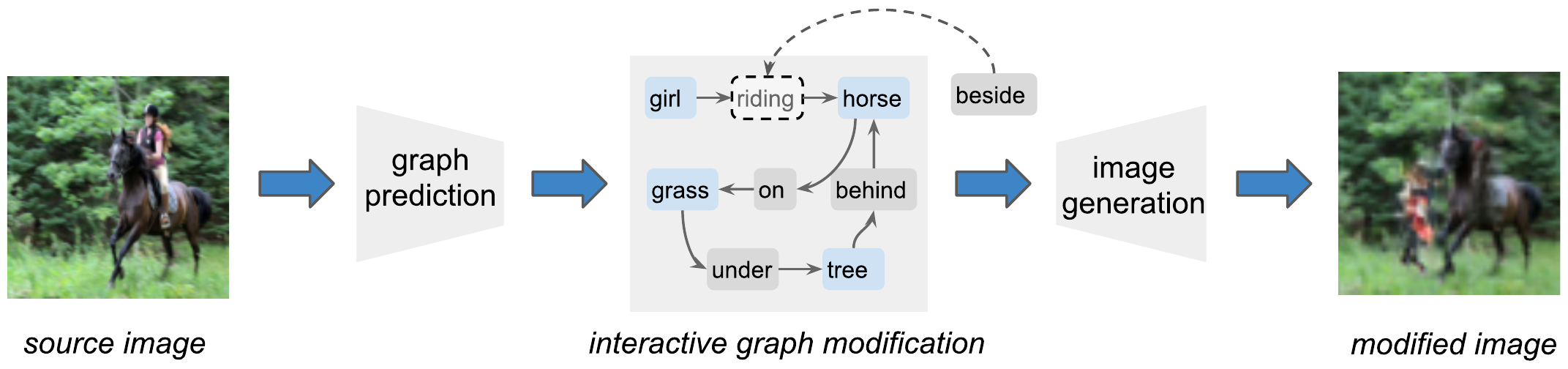}
\captionof{figure}{\textbf{Semantic Image Manipulation.} Given an image, we predict a semantic scene graph. The user interacts with the graph by making changes on the nodes and edges. Then, we generate a modified version of the source image, which respects the constellations in the modified graph.}
\label{fig:teaser}
\end{center}
\end{strip}
\begin{abstract}
Image manipulation can be considered a special case of image generation where the image to be produced is a modification of an existing image.
Image generation and manipulation have been, for the most part, tasks that operate on raw pixels. However, the remarkable progress in learning rich image and object representations has opened the way for tasks such as text-to-image or layout-to-image generation that are mainly driven by semantics. 
In our work, we address the novel problem of image manipulation from scene graphs, in which a user can edit images by merely applying changes in the nodes or edges of a semantic graph that is generated from the image. Our goal is to encode image information in a given constellation and from there on generate new constellations, such as replacing objects or even changing relationships between objects, while respecting the semantics and style from the original image. We introduce a spatio-semantic scene graph network that does not require direct supervision for constellation changes or image edits. This makes it possible to train the system from existing real-world datasets with no additional annotation effort. \blfootnote{Project page: \url{https://he-dhamo.github.io/SIMSG/}}
\end{abstract}

\input{1_introduction.tex}
\input{2_related_work.tex}
\input{3_method.tex}

\input{4_results.tex}

\input{5_conclusion.tex}

{\small
\bibliographystyle{ieee_fullname.bst}
\bibliography{references.bib}
}
\newpage

\input{6_supplement}

\end{document}

%% file: 1_introduction.tex
\section{Introduction}

The goal of image understanding is to extract rich and meaningful information from an image.  
Recent techniques based on deep representations are continuously pushing the boundaries of performance in recognizing objects~\citep{ren2015faster} and their relationships~\citep{lu2016visual} or producing image descriptions \citep{johnson2016densecap}.
Understanding is also necessary for image synthesis, \eg to generate natural looking images from an abstract semantic canvas~\cite{chen2017photographic,wang2018high,zhao2018image} or even from language descriptions~\cite{reed2016generative,zhang2018photographic,zhang2017stackgan,li2018storygan,hong2018inferring}.  
High-level image \textit{manipulation}, however, has received less attention. Image manipulation is still typically done at pixel level via photo editing software and low-level tools such as in-painting. Instances of higher-level manipulation are usually object-centric, such as facial modifications or reenactment.
A more abstract way of manipulating an image from its semantics, which includes objects, their relationships and attributes,
could make image editing easier with less manual effort from the user.

In this work, we present a method to perform semantic editing of an image by modifying a scene graph, which is a representation of the objects, attributes and interactions in the image (Figure \ref{fig:teaser}). 
As we show later, this formulation allows the user to choose among different editing functions.
For example, instead of manually segmenting, deleting and in-painting unwanted tourists in a holiday photo, the user can directly manipulate the scene graph and delete selected \texttt{<person>} nodes. 
Similarly, graph nodes can be easily replaced with different semantic categories, for example replacing \texttt{<clouds>} with \texttt{<sky>}. 
It is also possible to re-arrange the spatial composition of the image by swapping people or object nodes on the image canvas. 
To the best of our knowledge, this is the first approach to image editing that also enables semantic relationship changes, for example changing ``\textit{a person walking in front of the sunset}'' to ``\textit{a person jogging in front of the sunset}'' to create a more scenic image. 
The capability to reason and manipulate a scene graph is not only useful for photo editing. The field of robotics can also benefit from this kind of task, \eg a robot tasked to tidy up a room can\,---\,prior to acting\,---\,manipulate the scene graph of the perceived scene by moving objects to their designated spaces, changing their relationships and attributes: ``\textit{clothes lying on the floor}'' to ``\textit{folded clothes on a shelf}'', to obtain a realistic future view of the room.

Much previous work has focused either on generating a scene graph from an image \cite{newell2017pixels,li2018factorizable} or an image from a graph \cite{johnson2018image,ashual2019specifying}. Here we face challenges unique to the combined problem. For example, if the user changes a relationship attribute\,---\,\eg \texttt{<boy, sitting on, grass>} to \texttt{<boy, standing on, grass>}, the system needs to generate an image that contains the \emph{same} boy, thus preserving the identity as well as the content of the rest of the scene. 
Collecting a fully supervised data set, \ie a data set of ``before'' and ``after'' pairs together with the associated scene graph, poses major challenges. As we discuss below, this is not necessary. It is in fact possible to learn how to modify images using only training pairs of images and scenes graphs, which is data already available. 

In summary, we present a novel task; given an image, we manipulate it using the respective scene graph. Our contribution is a method to address this problem that does not require full supervision, \ie image pairs that contain scene changes. 
Our approach can be seen as semi-automatic, since the user does not need to manually edit the image but indirectly interacts with it through the nodes and edges of the graph. In this way, it is possible to make modifications with respect to visual entities in the image and the way they interact with each other, both spatially and semantically. 
Most prominently, we achieve various types of edits with a single model, including semantic relationship changes between objects. The resulting image preserves the original content, but allows the user to flexibly change and/or integrate new or modified content as desired.

%% file: 2_related_work.tex
\section{Related Work}

\begin{figure*}[t]
\centering
\includegraphics[width=\linewidth]{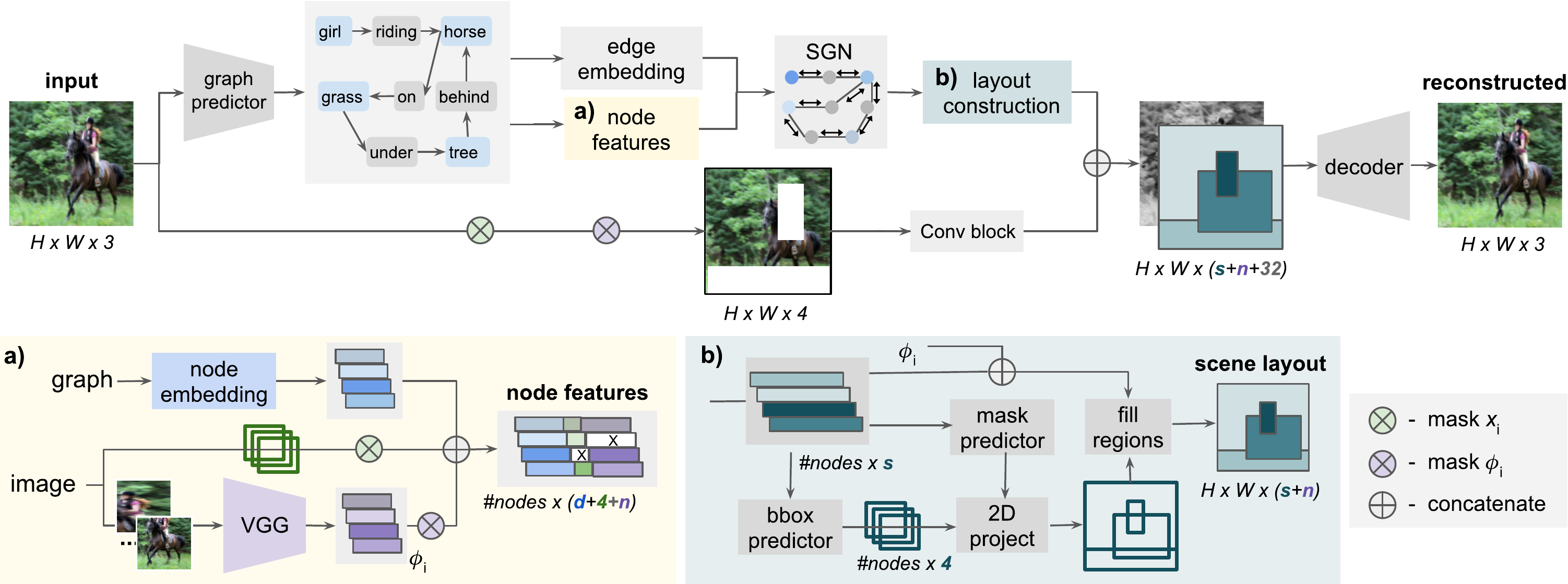}
\caption{\textbf{Overview of the training strategy.} \emph{Top:} Given an image, we predict its scene graph and reconstruct the input from a masked representation. a) The graph nodes $o_i$ \emph{(blue)} are enriched with bounding boxes $x_i$ \emph{(green)} and visual features $\phi_i$ \emph{(violet)} from cropped objects. We randomly mask boxes $x_i$, object visual features $\phi_i$ and the source image; the model then reconstructs the same graph and image utilizing the remaining information. b) The per-node feature vectors are projected to 2D space, using the bounding box predictions from SGN.}
\label{fig:method}
\end{figure*}

\paragraph{Conditional image generation}
The success of deep generative models~\cite{goodfellow2014generative,kingma2013auto,radford2015unsupervised,van2016pixel,van2016conditional} has significantly contributed to advances in (un)conditional image synthesis. 
Conditional image generation methods model the conditional distribution of images given some prior information. 
For example, several practical tasks such as denoising or inpainting can be seen as generation from noisy or partial input. 
Conditional models have been studied in literature for a variety of use cases, conditioning the generation process on image labels~\cite{mirza2014conditional,odena2017conditional}, attributes~\cite{yan2016attribute2image}, lower resolution images~\cite{ledig2017photo}, semantic segmentation maps~\cite{chen2017photographic,wang2018high}, natural language descriptions~\cite{reed2016generative,zhang2018photographic,zhang2017stackgan,li2018storygan} or generally translating from one image domain to another using paired~\cite{isola2017image} or unpaired data~\cite{zhu2017unpaired}. 
Most relevant to our approach are methods that generate natural scenes from layout~\cite{hong2018inferring,zhao2018image} or scene graphs~\cite{johnson2018image}.

\paragraph{Image manipulation}
Unconditional image synthesis is still an open challenge when it comes to complex scenes. Image manipulation, on the other hand, focuses on image parts in a more constrained way that allows to generate better quality samples. 
Image manipulation based on semantics has been mostly restricted to object-centric scenarios; for example, editing faces automatically using attributes~\cite{choi2018stargan,lample2017fader,zhao2018modular} or via manual edits with a paintbrush and scribbles~\cite{brock2016neural,zhu2016generative}. 
Also related is image composition which also makes use of individual objects~\citep{azadi2018compositional} and faces the challenge of decoupling appearance and geometry~\citep{zhan2018spatial}.

On the level of scenes, the most common examples based on generative models are inpainting~\cite{pathak2016context}, in particular conditioned on semantics~\cite{yeh2017semantic} or user-specified contents~\cite{Zhao2018GuidedII,jo2019sc}, as well as object removal~\citep{shetty2018adversarial,Dhamo2019iccv}. %
Image generation from semantics also supports interactive editing by applying changes to the semantic map~\cite{wang2018high}. Differently, we follow a semi-automatic approach to address all these scenarios using a single general-purpose model and incorporating edits by means of a scene graph. On another line, Hu \etal~\cite{Hu:2013} propose a hand-crafted image editing approach, which uses graphs to carry out library-driven replacement of image patches. While \cite{Hu:2013} focuses on copy-paste tasks, our framework allows for high-level semantic edits and deals with object deformations. 

Our method is trained by reconstructing the input image so it does not require paired data. A similar idea is explored by Yao \etal~\cite{yao20183d} for 3D-aware modification of a scene (\ie 3D object pose) by disentangling semantics and geometry. 
However, this approach is limited to a specific type of scenes (streets) and target objects (cars) and requires CAD models.
Instead, our approach addresses semantic changes of objects and their relationships in natural scenes, which is made possible using scene graphs.  

\paragraph{Images and scene graphs}

Scene graphs provide abstract, structured representations of image content. Johnson \etal~\cite{johnson15} first defined a scene graph as a directed graph representation that contains objects and their attributes and relationships, \ie how they interact with each other. 
Following this graph representation paradigm, different methods have been proposed to generate scene graphs from images~\cite{xu2017scenegraph,herzig2018mapping,qi2018attentive,zellers2018neural,li2017scene,yang2018graph,li2018factorizable,newell2017pixels}. 
By definition, scene graph generation mainly relies on successfully detecting visual entities in the image (object detection)~\citep{ren2015faster} and recognizing how these entities interact with each other (visual relationship detection)~\citep{dai2017detecting,jae2018tensorize,lu2016visual,sadeghi2011recognition,yin2018zoom}.

The reverse and under-constrained problem is to generate an image from its scene graph, which has been recently addressed by Johnson~\etal using a graph convolution network (GCN) to decode the graph into a layout and consecutively translate it into image~\citep{johnson2018image}. We build on this architecture and propose additional mechanisms for information transfer from an image that act as conditioning for the system, when the goal is image editing and not free-form generation. 
Also related is image generation directly from layouts~\citep{zhao2018image}. Very recent related work focuses on interactive image generation from scene graphs \cite{ashual2019specifying} or layout \cite{Sun_2019_ICCV}. 
These methods differ from ours in two aspects. First, while \cite{ashual2019specifying,Sun_2019_ICCV} process a graph/layout to generate multiple variants of an image, our method manipulates an \emph{existing} image. 
Second, we present complex semantic relationship editing, while they use graphs with simplified spatial relations\,---\,\eg relative object positions such as \texttt{left of} or \texttt{above} in \cite{ashual2019specifying}\,---\,or without relations at all, as is the case for the layout-only approach in \cite{Sun_2019_ICCV}. %

%% file: 3_method.tex
\section{Method}
\label{sec:method}
The focus of this work is to perform semantic manipulation of images without direct supervision for image edits, \ie without paired data of original and modified content. 
Starting from an input image $I$, we generate its scene graph $\mathcal{G}$ that serves as the means of interaction with a user. 
We then generate a new image $I'$ from the user-modified graph representation $\tilde{\mathcal{G}}$ and the original content of $I$. An overview of the method is shown in Figure \ref{fig:teaser}. 
Our method can be split into three interconnected parts. 
The first step is scene graph generation, where we encode the image contents in a spatio-semantic scene graph, designed so that it can easily be manipulated by a user. 
Second, during inference, the user manipulates the scene graph by modifying object categories, locations or relations by directly acting on the nodes and edges of the graph. 
Finally, the output image is generated from the modified graph. 
Figure \ref{fig:method} shows the three components and how they are connected. 

A particular challenge in this problem is the difficulty in obtaining training data, \ie matching pairs of source and target images together with their corresponding scene graphs. 
To overcome these limitations, we demonstrate a method that learns the task by image reconstruction in an unsupervised way. %
Due to readily available training data, graph prediction instead is learned with full supervision. 

\subsection{Graph Generation}
Generating a scene graph from an image is a well-researched problem \cite{xu2017scenegraph,zellers2018neural,newell2017pixels,li2018factorizable} 
and amounts to describing the image with a directed graph $\mathcal{G} = (\mathcal{O}, \mathcal{R})$ of objects $\mathcal{O}$ (nodes) and their relations $\mathcal{R}$ (edges).    
We use a state-of-the-art method for scene graph prediction (F-Net)~\cite{li2018factorizable} and build on its output.   
Since the output of the system is a generated image, our goal is to encode as much image information in the scene graph as possible --- additional to semantic relationships. 
We thus define objects as triplets $o_i = (c_i, \phi_i, x_i) \in \mathcal{O}$, where $c_i \in \mathbb{R}^d$ is a $d$-dimensional, learned embedding of the $i$-th object category and $x_i \in \mathbb{R}^4$ represents the four values defining the object's bounding box.
$\phi_i\in \mathbb{R}^n$ is a visual feature encoding of the object which can be obtained from a convolutional neural network (CNN) pre-trained for image classification. 
Analogously, for a given relationship between two objects $i$ and $j$, we learn an embedding $\rho_{ij}$ of the relation class $r_{ij}\in\mathcal{R}$.

One can also see this graph representation as an augmentation of a simple graph---that only contains object and predicate categories---with image features and spatial locations. 
Our graph contains sufficient information to preserve the identity and appearance of objects even when the corresponding locations and/or relationships are modified. 

\subsection{Spatio-semantic Scene Graph Network}
At the heart of our method lies the spatio-semantic scene graph network (SGN) that operates on the (user-) modified graph. 
The network learns a graph transformation that allows information to flow between objects, along their relationships. 
The task of the SGN is to learn robust object representations that will be then used to reconstruct the image. 
This is done by a series of convolutional operations on the graph structure.

The graph convolutions are implemented by an operation $\tau_e$ on edges of the graph
\begin{equation}
    (\alpha_{ij}^{(t+1)}, \rho_{ij}^{(t+1)}, \beta_{ij}^{(t+1)}) = \tau_e\left(
    \nu_i^{(t)}, \,\rho_{ij}^{(t)}, \,\nu_j^{(t)}
    \right) ,
\end{equation}
with $\nu_i^{(0)} = o_i$, where $t$ represents the layer of the SGN and $\tau_e$ is implemented as a multi-layer perceptron (MLP).
Since nodes can appear in several edges, the new node feature $\nu_i^{(t+1)}$ is computed by averaging the results from the edge-wise transformation, followed by another projection $\tau_n$
\begin{equation}
\scalemath{0.92}{
    \nu_i^{(t+1)} = \tau_n \Bigg( \frac{1}{N_i} \bigg(
    \sum_{j | (i,j) \in \mathcal{R}} \alpha_{ij}^{(t+1)} + 
    \sum_{k | (k,i) \in \mathcal{R}} \beta_{ki}^{(t+1)} \bigg) \Bigg)
    }
\end{equation}
where $N_i$ represents the number of edges that start or end in node $i$. After $T$ graph convolutional layers, the last layer predicts one latent representation per node, \ie per object. This output object representation consists of predicted bounding box coordinates $\hat{x}_i \in \mathbb{R}^4$, a spatial binary mask $\hat{m}_i \in \mathbb{R}^{M \times M}$  and a node feature vector $\psi_i \in \mathbb{R}^s$. 
Predicting coordinates for each object is a form of reconstruction, since object locations are known and are already encoded in the input $o_i$. As we show later, this is needed when modifying the graph, for example for a new node to be added.    
The predicted object representation will be then reassembled into the spatial configuration of an image, as the scene layout. 

\subsection{Scene Layout}
The next component is responsible for transforming the graph-structured representations predicted by the SGN back into a 2D spatial arrangement of features, which can then be decoded into an image. 
To this end, we use the predicted bounding box coordinates $\hat{x}_i$ to project the masks $\hat{m}_i$ in the proper region of a 2D representation of the same resolution as the input image. 
We concatenate the original visual feature $\phi_i$ with the node features $\psi_i$ to obtain a final node feature.
The projected mask region is then filled with the respective features, while the remaining area is padded with zeros. This process is repeated for all objects, resulting in $|\mathcal{O}|$ tensors of dimensions $(n+s) \times H \times W$, which are aggregated through summation into a single layout for the image. 
The output of this component is an intermediate representation of the scene, which is rich enough to reconstruct an image.     

\subsection{Image Synthesis}
The last part of the pipeline is the task of synthesizing a target image from the information in the source image $I$ and the layout prediction. For this task, we employ two different decoder architectures, cascaded refinement networks (CRN)~\citep{chen2017photographic} (similar to \citep{johnson2018image}), as well as SPADE \cite{park2019SPADE}, originally proposed for image synthesis from a semantic segmentation map.
We condition the image synthesis on the source image by concatenating the predicted layout with extracted low-level features from the source image. %
In practice, prior to feature extraction, regions of $I$ are occluded using a mechanism explained in Section \ref{sec:train}. We fill these regions with Gaussian noise to introduce stochasticity for the generator.

\subsection{Training}
\label{sec:train}
Training the model with full supervision would require annotations in the form of quadruplets $(I, \mathcal{G}, \mathcal{G}', I')$ where an image $I$ is annotated with a scene graph $\mathcal{G}$, a modified graph $\mathcal{G}'$ and the resulting modified image $I'$.
Since acquiring ground truth $(I', \mathcal{G}')$ is difficult, our goal is to train a model supervised only by $(I, \mathcal{G})$ through reconstruction. Thus, we generate annotation quadruplets $(\tilde{I}, \tilde{\mathcal{G}}, \mathcal{G}, I)$ using the available data $(I, \mathcal{G})$ as the \emph{target} supervision and simulate $(\tilde{I}, \tilde{\mathcal{G}})$ via a random masking procedure that operates on object instances. 
During training, an object's visual features $\phi_i$ are masked with probability $p_{\phi}$. Independently, we mask the bounding box $x_i$ with probability $p_x$. 
When ``hiding'' input information, image regions corresponding to the hidden nodes are also occluded prior to feature extraction. 

Effectively, this masking mechanism transforms the editing task into a reconstruction problem. 
At run time, a real user can directly edit the nodes or edges of the scene graph.
Given the edit, the image regions subject to modification are occluded, and the network, having learned to reconstruct the image from the scene graph, will create a plausible modified image. 
Consider the example of a person riding a horse (Figure~\ref{fig:teaser}). 
The user wishes to apply a change in the way the two entities interact, modifying the predicate from \texttt{riding} to \texttt{beside}. 
Since we expect the spatial arrangement to change, we also discard the localization $x_i$ of these entities in the original image; their new positions $\hat{x}_i$ will be estimated given the layout of the rest of the scene (\eg grass, trees). To encourage this change, the system should automatically mask the original image regions related to the target objects. However, to ensure that the \emph{visual identities} of horse and rider are preserved through the change, their visual feature encodings $\phi_i$ must remain unchanged. 

We use a combination of loss terms to train the model. 
The bounding box prediction is trained by minimizing the $L_1$-norm: $\mathcal{L}_b = \left\lVert x_i - \hat{x}_i \right\rVert_1^1$, with weighting term $\lambda_b$.
The image generation task is learned by adversarial training with two discriminators. A local discriminator $D_\text{obj}$ operates on each reconstructed region to ensure that the generated patches look realistic. We also apply an auxiliary classifier loss~\cite{Odena2017ConditionalIS} to ensure that $D_\text{obj}$ is able to classify the generated objects into their real labels. A global discriminator $D_\text{global}$ encourages consistency over the entire image.
Finally, we apply a photometric loss term $\mathcal{L}_{r}=\|I - I'\|_1$ to enforce the image content to stay the same in regions that are not subject to change. The total synthesis loss is then
\begin{equation}
  \begin{split}
      \mathcal{L}_\text{synthesis} & =  \mathcal{L}_r + \lambda_g \min_G \max_D \mathcal{L}_\text{GAN,global} \\  &\quad + \lambda_o \min_G \max_D \mathcal{L}_\text{GAN,obj} + \lambda_a \mathcal{L}_\text{aux,obj},
  \end{split}
\end{equation}
where $\lambda_g$, $\lambda_o$, $\lambda_a$ are weighting factors and
\begin{equation}
  \mathcal{L}_\text{GAN} = \mathop{\mathbb{E}}_{q\sim p_{\textrm{real}}} \log D(q) + \mathop{\mathbb{E}}_{q\sim p_{\textrm{fake}}} \log(1 - D(q)),
\end{equation}
where $p_{real}$ corresponds to the ground truth distribution (of each object or the whole image) and $p_{fake}$ is the distribution of generated (edited) images or objects, while $q$ is the input to the discriminator which is sampled from the real or fake distributions. When using SPADE, we additionally employ a perceptual loss term $\lambda_p \mathcal{L}_p$ and a GAN feature loss term $\lambda_f \mathcal{L}_f$ following the original implementation \cite{park2019SPADE}. Moreover, $D_{global}$ becomes a multi-scale discriminator.  

Full implementation details regarding the architectures, hyper-parameters and training can be found in the Appendix.

%% file: 4_results.tex
\section{Experiments}

We evaluate our method quantitatively and qualitatively on two datasets, CLEVR~\citep{johnson2017clevr} and Visual Genome~\citep{krishna2017visual}, with two different motivations. As CLEVR is a synthetic dataset, obtaining ground truth pairs for image editing is possible, which allows quantitative evaluation of our method. On the other hand, experiments on Visual Genome (VG) show the performance of our method in a real, much less constrained, scenario. 
In absence of source-target image pairs in VG, we evaluate an image in-painting proxy task and compare to a baseline based on sg2im \citep{johnson2018image}. 
We report results for standard image reconstruction metrics: the structural similarity index (SSIM), %
mean absolute error (MAE) and perceptual error (LPIPS) \cite{zhang2018perceptual}. To assess the image generation quality and diversity, we report the commonly used inception score (IS) \cite{Salimans2016ImprovedTF} and the FID \cite{Heusel2017GANsTB} metric. %

\paragraph{Conditional sg2im baseline (Cond-sg2im).} 
We modify the model of~\citep{johnson2018image} to serve as a baseline. Since their method generates images directly from scene graphs without a source image, we condition their image synthesis network on the input image by concatenating it with the layout component (instead of noise in the original work). To be comparable to our approach, we mask image regions corresponding to the target objects prior to concatenation. 

\paragraph{Modification types.}
Since image editing using scene graphs is a novel task, we define several modification modes, depending on how the user interacts with the graph. 
\textbf{Object removal:} A node is removed entirely from the graph together with all the edges that connect this object with others. The source image region corresponding to the object is occluded.
\textbf{Object replacement:} A node is assigned to a different semantic category. We do not remove the full node; however, the visual encoding $\phi_i$ of the original object is set to zero, as it does not describe the novel object. The location of the original entity is used to keep the new object in place, while size comes from the bounding box estimated from the SGN, to fit the new category.
\textbf{Relationship change:} This operation usually involves re-positioning of entities. The goal is to keep the subject and object but change their interaction, \eg \texttt{<sitting>} to \texttt{<standing>}.
Both the original and novel appearance image regions are occluded, to enable background in-painting and target object generation.
The visual encodings $\phi_i$ are used to condition the SGN and maintain the visual identities of objects on re-appearance.  

\begin{figure*}[t]
\centering
\includegraphics[width=0.95\linewidth]{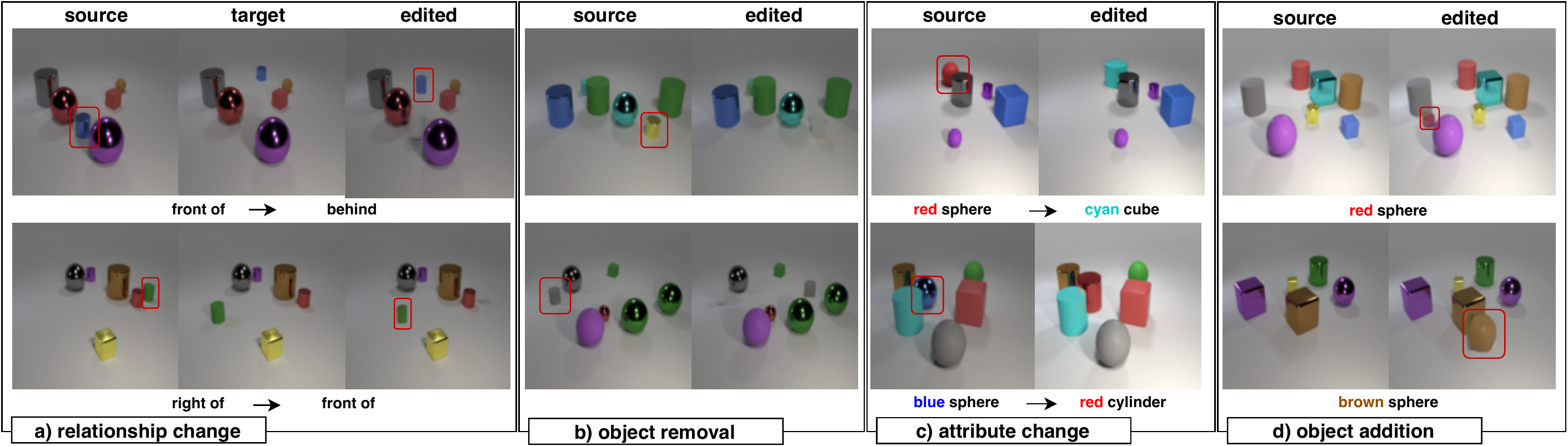}
\caption{\textbf{Image manipulation on CLEVR} We compare different changes in the scene including changing the relationship between two objects, node removal and changing a node (corresponding to attribute changing).} 
\label{fig:clevr_modes}
\end{figure*}

\begin{table}[t]
\centering
\small
\resizebox{\columnwidth}{!}{
\begin{tabular}{l@{\hskip 7pt} c @{\hskip 7pt} c@{\hskip 7pt} c @{\hskip 7pt} c@{\hskip 7pt} c@{\hskip 7pt} c}
\toprule
 \multirow{2}{*}{Method} & \multicolumn{4}{c}{All pixels} & \multicolumn{2}{c}{RoI only} \\
 \cmidrule(l{1pt}r{9pt}){2-5} \cmidrule(l{1pt}r{3pt}){6-7}
 & \footnotesize{MAE $\downarrow$} & \footnotesize{SSIM $\uparrow$} & \footnotesize{LPIPS $\downarrow$} & \footnotesize{FID $\downarrow$} & \footnotesize{MAE $\downarrow$} & \footnotesize{SSIM $\uparrow$}\\ %
\midrule
Full-sup & $6.75$ & $\bm{97.07}$ & $0.035$ & $\bm{3.35}$ & $9.34$ & $93.49$\\
Ours (CRN) & $7.83$ & $96.16$ & $0.036$ & $6.32$ & $10.09$ & $93.54$\\
Ours (SPADE) & $\bm{5.47}$ & $96.51$ & $0.035$ & $4.73$ & $\bm{7.22}$ & $\bm{94.98}$ \\
\bottomrule
\end{tabular}}
\caption{\textbf{Image manipulation on CLEVR.} We compare our method with a fully-supervised baseline. Detailed results for all modification types are reported in the Appendix.}
\label{tab:clevr1}
\end{table}

\subsection{Synthetic Data}
We use the CLEVR framework~\citep{johnson2017clevr} to generate a dataset (for details please see the Appendix) of image and scene graph editing pairs $(I, \mathcal{G}, \mathcal{G}', I')$, to evaluate our method with exact ground truth.  

We train our model \textit{without making use of image pairs} and compare our approach to a fully-supervised setting. %
When training with full supervision the complete source image and target graph are given to the model and the model is trained by minimizing the $\mathcal{L}_1$ loss to the ground truth target image instead of the proposed masking scheme. 

Table~\ref{tab:clevr1} reports the mean SSIM, MAE, LPIPS and FID on CLEVR for the manipulation task (replacement, removal, relationship change and addition). Our method performs better or on par with the fully-supervised setting, on the reconstruction metrics, which shows the capability of synthesizing meaningful changes. The FID results suggest that additional supervision for pairs, if available, would lead to improvement in the visual quality.
Figure~\ref{fig:clevr_modes} shows qualitative results of our model on CLEVR. At test time, we apply changes to the scene graph in four different modes: changing relationships (a), removing an object (b), adding an object (d) or changing its identity (c). We highlight the modification with a bounding box drawn around the selected object. 

\subsection{Real Images}

\begin{figure}[t]
\centering
\includegraphics[width=\linewidth]{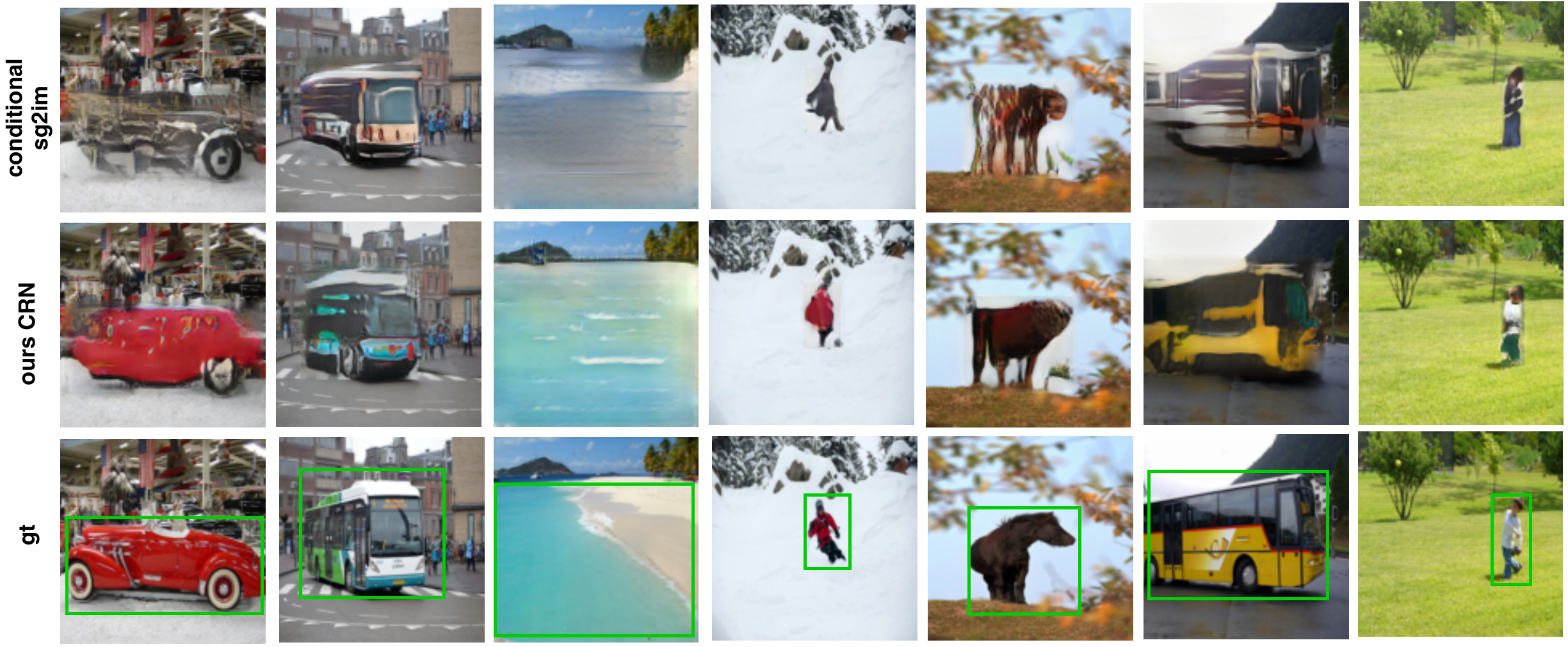}
\caption{\textbf{Visual feature encoding.} Comparison between the baseline (top) and our method (center). The scene graph remains unchanged; an object in the image is occluded, while $\phi_i$ and $x_i$ are active. Our latent features $\phi_i$ preserve appearance when the objects are masked from the image. } 
\label{fig:auto_comparison}
\end{figure}

\begin{table*}[t]
    \centering
    \footnotesize %
    \begin{tabular}{l@{\hskip 8pt} l@{\hskip 16pt} c@{\hskip 26pt} c@{\hskip 12pt} c@{\hskip 12pt} c@{\hskip 12pt} c@{\hskip 12pt} c@{\hskip 12pt} c@{\hskip 12pt} c@{\hskip 10pt} c@{\hskip 10pt}}
    \toprule
    \multicolumn{2}{l}{\multirow{2}{*}{Method}} & \multirow{2}{*}{Decoder} & \multicolumn{5}{c}{All pixels} & \multicolumn{2}{c}{RoI only} \\
    \cmidrule(l{1pt}r{9pt}){4-8} \cmidrule(l{1pt}r{3pt}){9-10}
     & & & MAE $\downarrow$ & SSIM $\uparrow$ & LPIPS $\downarrow$ & FID $\downarrow$ & IS $\uparrow$ & MAE $\downarrow$ & SSIM $\uparrow$ \\
    \midrule
    \multicolumn{2}{l}{ISG~\citep{ashual2019specifying} (Generative, GT)}  & Pix2pixHD & $46.44$ & $28.10$ & $0.32$ & 58.73 & $6.64$\xpm{0.07} & - & - \\
    \multicolumn{2}{l}{Ours (Generative, GT)} & CRN & $41.57$ & $33.9$ & $0.34$ & $89.55$ & $6.03$\xpm{0.17} & - & - \\
    \multicolumn{2}{l}{Ours (Generative, GT)} & SPADE & $41.88$ & $34.89$ & $0.27$ & $44.27$ & $7.86$\xpm{0.49} & - & - \\
    \midrule \midrule
    \multicolumn{2}{l}{Cond-sg2im \citep{johnson2018image} (GT)} & CRN & $14.25$ & $84.42$ & $0.081$ & 13.40 & $11.14$\xpm{0.80} & $29.05$ & $52.51$\\
    Ours (GT) & w/o $\phi_i$ & CRN &  $9.83$ & $86.52$ & $0.073$ & $10.62$ & $11.45$\xpm{0.61} & $27.16$ & $52.01$ \\
    Ours (GT) & w/ $\phi_i$ & CRN &  $\bm{7.43}$ & $\bm{88.29}$ & $0.058$ & $11.03$ & $11.22$\xpm{0.52} & $\bm{20.37}$ & $\bm{60.03}$\\
    Ours (GT) & w/o $\phi_i$ & SPADE &  $10.36$ & $86.67$ & $0.069$ & $8.09$ & $12.05$\xpm{0.80} & $27.10$ & $54.38$ \\
    Ours (GT) & w/ $\phi_i$ & SPADE &  $8.53$ & $87.57$ & $\bm{0.051}$ & $\bm{7.54}$ & $\bm{12.07}$\xpm{0.97} & $21.56$ & $58.60$\\
    \midrule
    Ours (P) & w/o $\phi_i$ & CRN & $9.24$ & $87.01$ & 0.075 & 18.09 & $10.67$\xpm{0.43} & $29.08$ & $48.62$\\
    Ours (P) & w/ $\phi_i$ & CRN & $7.62$ & $88.31$ & 0.063 & 19.49 & $10.18$\xpm{0.27} & $22.89$ & $55.07$ \\
    Ours (P) & w/o $\phi_i$ & SPADE & $13.16$ & $84.61$ & 0.083 & 16.12 & $10.45$\xpm{0.15} & $32.24$ & $47.25$\\
    Ours (P) & w/ $\phi_i$ & SPADE & $13.82$ & $83.98$ & $0.077$ & $16.69$ & $10.61$\xpm{0.37} & $28.82$ & $49.34$ \\
    \bottomrule
\end{tabular}%
    \caption{\textbf{Image reconstruction on Visual Genome.} We report the results using ground truth scene graphs (GT) and predicted scene graphs (P). (Generative) indicates experiments in full generative setting, \ie the whole input image is masked out.}
    \label{tab:vg}
\end{table*}

We evaluate our method on Visual Genome~\citep{krishna2017visual} to show its performance on natural images. Since there is no ground truth for modifications, we formulate the quantitative evaluation as image reconstruction. In this case, objects are occluded from the original image and we measure the quality of the reconstruction. The qualitative results better illustrate the full potential of our method.   %

\paragraph{Feature encoding.}
First, we quantify the role of the visual feature $\phi_i$ in encoding visual appearance. 
For a given image and its graph, we use all the associated object locations $x_i$ and visual features (w/ $\phi_i$) to condition the SGN. However, the region of the conditioning image corresponding to a candidate node is masked. The task can be interpreted as conditional in-painting. 
We test our approach in two scenarios; using ground truth graphs (GT) and graphs predicted from the input images (P). 
We evaluate over all objects in the test set and report the results in Table~\ref{tab:vg}, measuring the reconstruction error a) over all pixels and b) in the target area only (RoI).
We compare to the same model without using visual features (w/o $\phi_i$) but only the object category to condition the SGN. %
Naturally, in all cases, including the missing region's visual features improves the reconstruction metrics (MAE, SSIM, LPIPS). In contrast, inception score and FID remain similar, as these metrics do not consider similarity between direct corresponding pairs of generated and ground truth images. From Table~\ref{tab:vg} one can observe that while both decoders perform similarly in reconstruction metrics (CRN is slightly better), SPADE dominates for the FID and inception score, indicating higher visual quality. 

\begin{figure*}[!ht]
\centering
\includegraphics[width=\linewidth]{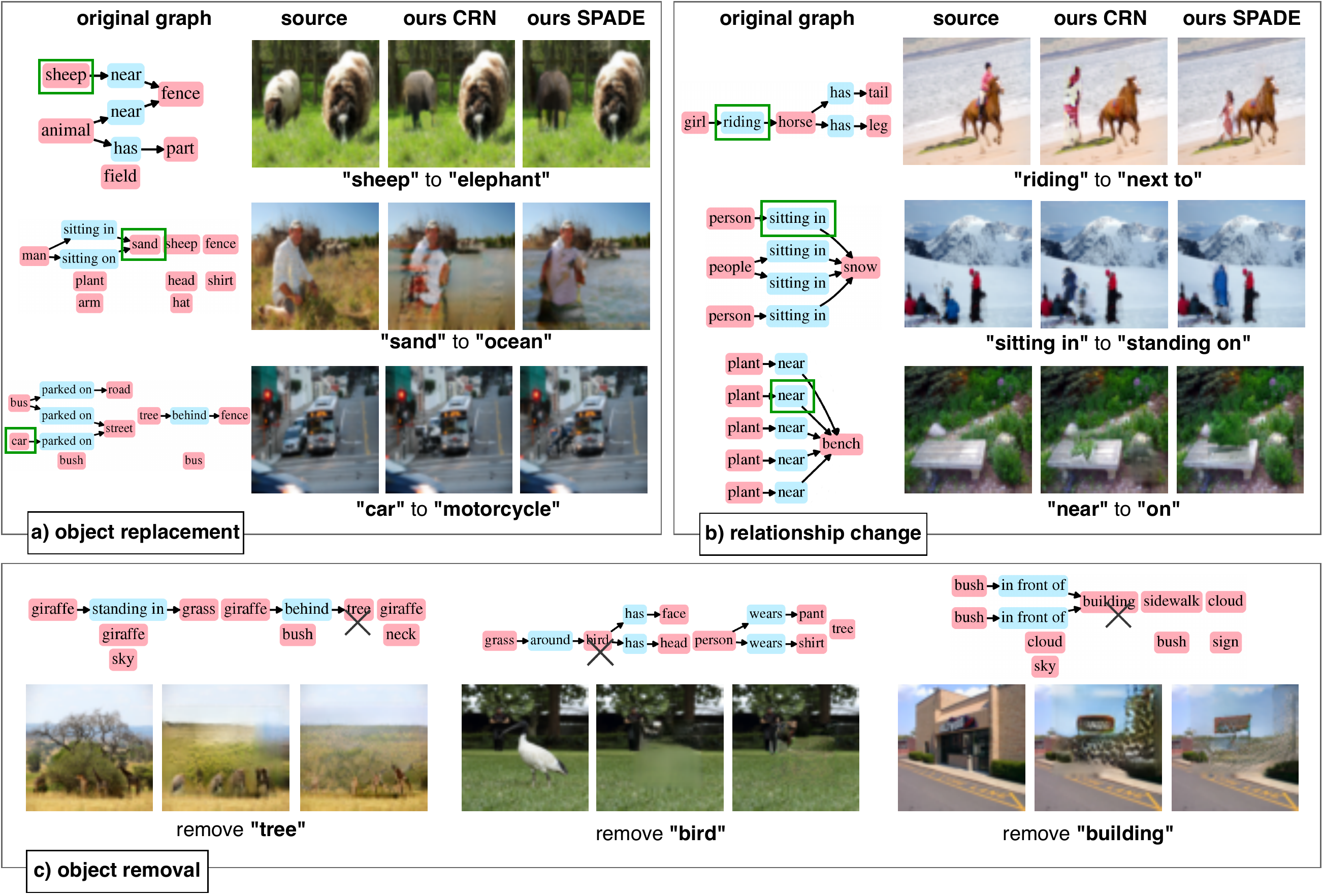}
\caption{\textbf{Image manipulation} Given the source image and the GT scene graph, we semantically edit the image by changing the graph. \textbf{a)} object replacement, \textbf{b)} relationship changes, \textbf{c)} object removal. Green box indicates the changed node or edge.}
\label{fig:modification_vg}
\end{figure*}

\begin{figure*}[h]
\centering
\includegraphics[width=0.92\linewidth]{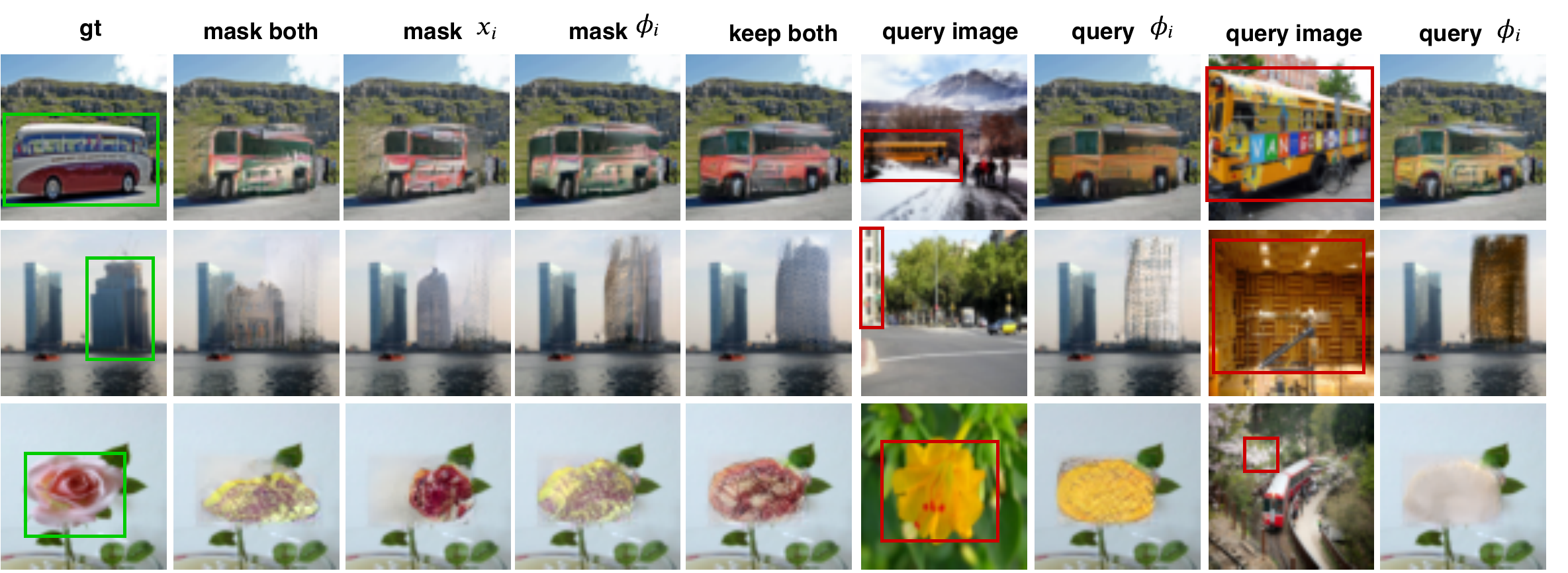}
\caption{\textbf{Ablation of the method components} We present all the different combinations in which the method operates - \ie masked vs. active bounding boxes $x_i$ and/or visual features $\phi_i$. When using a query image, we extract visual features of the object annotated with a red bounding box and update the node of an object of the same category in the original image.} 
\label{fig:ablation}
\end{figure*}

To evaluate our method in a fully generative setting, we mask the whole image and only use the encoded features $\phi_i$ for each object. We compare against the state of the art in interactive scene generation (ISG)~\cite{ashual2019specifying}, evaluated in the same setting. Since our main focus is on semantically rich relations, we trained \cite{ashual2019specifying} on Visual Genome, utilizing their publicly available code. Table~\ref{tab:vg} shows comparable reconstruction errors for the generative task, while we clearly outperform \cite{ashual2019specifying} when a source image is given. This motivates our choice of directly manipulating an existing image, rather than fusing different node features, as parts of the image need to be preserved. Inception score and FID mostly depend on the decoder architecture, where SPADE outperforms Pix2pixHD and CRN. 

Figure \ref{fig:auto_comparison} illustrates qualitative examples.  It can be seen that both our method and the cond-sg2im baseline, generate plausible object categories and shapes. However, with our approach, visual features from the original image can be successfully transferred to the output. In practice, this property is particularly useful when we want to re-position objects in the image without changing their identity. 

\paragraph{Main task: image editing.}
We illustrate visual results in three different settings in Figure \ref{fig:modification_vg}\,---\,object removal, replacement and relationship changes. All image modifications are made by the user at test time, by changing nodes or edges in the graph. %
We show diverse replacements (a), from small objects to background components. The novel entity adapts to the image context, \eg the ocean (second row) does not occlude the person, which we would expect in standard image inpainting. 
A more challenging scenario is to change the way two objects interact, which typically involves re-positioning. %
Figure \ref{fig:modification_vg} (b) shows that the model can differentiate between semantic concepts, such as \texttt{sitting} vs. \texttt{standing} and \texttt{riding} vs. \texttt{next to}. The objects are rearranged meaningfully according to the change in relationship type. In the case of object removal (c), the method performs well for backgrounds with uniform texture, but can also handle more complex structures, such as the background in the first example. Interestingly, when the building on the rightmost example is removed, the remaining sign is improvised standing in the bush. More results are shown in the Appendix. 

\paragraph{Component ablation.} In Figure \ref{fig:ablation} we qualitatively ablate the components of our method. For a certain image, we mask out a certain object instance which we aim to reconstruct. We test the method under all the possible combinations of masking bounding boxes $x_i$ and/or visual features $\phi_i$ from the augmented graph representation. Since it might be of interest to in-paint the region with a different object (changing either the category or style), we also experiment with an additional setting, in which external visual features $\phi$ are extracted from an image of the query object. Intuitively, masking the box properties leads to a small shift in the location and size of the reconstructed object, while masking the object features can result in an object with a different identity than that in the original image.

%% file: 5_conclusion.tex
\section{Conclusion}

We have presented a novel task\,---\,semantic image manipulation using scene graphs\,---\,and have shown a novel approach to tackle the learning problem in a way that does not require training pairs of original and modified image content. The resulting system provides a way to change both the content and relationships among scene entities by directly interacting with the nodes and edges of the scene graph.
We have shown that the resulting system is competitive with baselines built from existing image synthesis methods, and qualitatively provides compelling evidence for its ability to support modification of real-world images. Future work will be devoted to further enhancing these results, and applying them to both interactive editing and robotics applications. 

\vspace{-0.5em}
\paragraph{Acknowledgements}
We gratefully acknowledge the Deutsche Forschungsgemeinschaft (DFG) for supporting this research work, under the project $\#381855581$. Christian Rupprecht is supported by ERC IDIU-638009.

%% file: 6_supplement.tex
\section{Appendix}
In the following, we provide additional results, as well as full details about the implementation and training of our method.  
Code and data splits for future benchmarks will be released in the project web-page\footnote{\url{https://he-dhamo.github.io/SIMSG/}}.

\subsection{More Qualitative Results}
\begin{figure*}[h]
\centering
\includegraphics[width=0.8\linewidth]{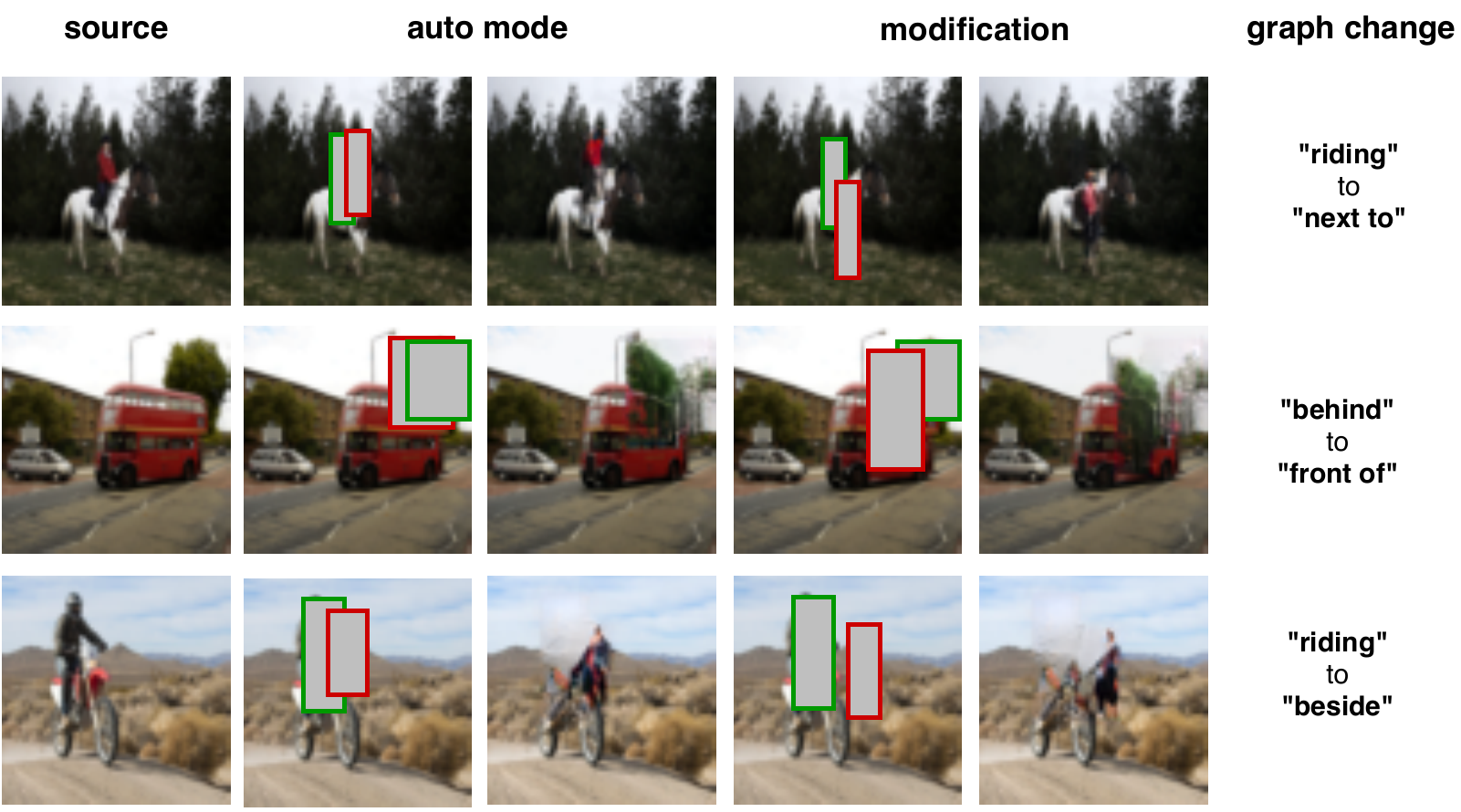}
\caption{\textbf{Re-positioning tested in more detail.} We mask the bounding box $x_i$ of an object and generate a target image in two modes. We choose a relationship that involves this object. In auto-mode (left) the relationship is kept unchanged. In modification mode, we change the relationship. Red: Predicted box for the auto-encoded or altered setting. Green: ground truth bounding box for the original relationship.} 
\label{fig:reposition_obj}
\end{figure*}

\paragraph{Relationship changes} Figure \ref{fig:reposition_obj} illustrates in more detail our method's behavior during relationship changes.
We investigate how the bounding box placement and the image generation of an object changes when one of its relationships is altered. 
We compare results between auto-encoding mode and modification mode. 
The bounding box coordinates are masked in both cases so that the model can decide where to position the target object depending on the relationships. 
In auto-encoding mode, the predicted boxes (red) end up in a valid location for the original relationship, while in the altered setup, the predicted boxes respect the changed relationship, \eg in auto mode, the person remains on the horse, while in modification mode the box moves beside the horse. 

\begin{figure*}
\centering
\includegraphics[width=0.62\linewidth]{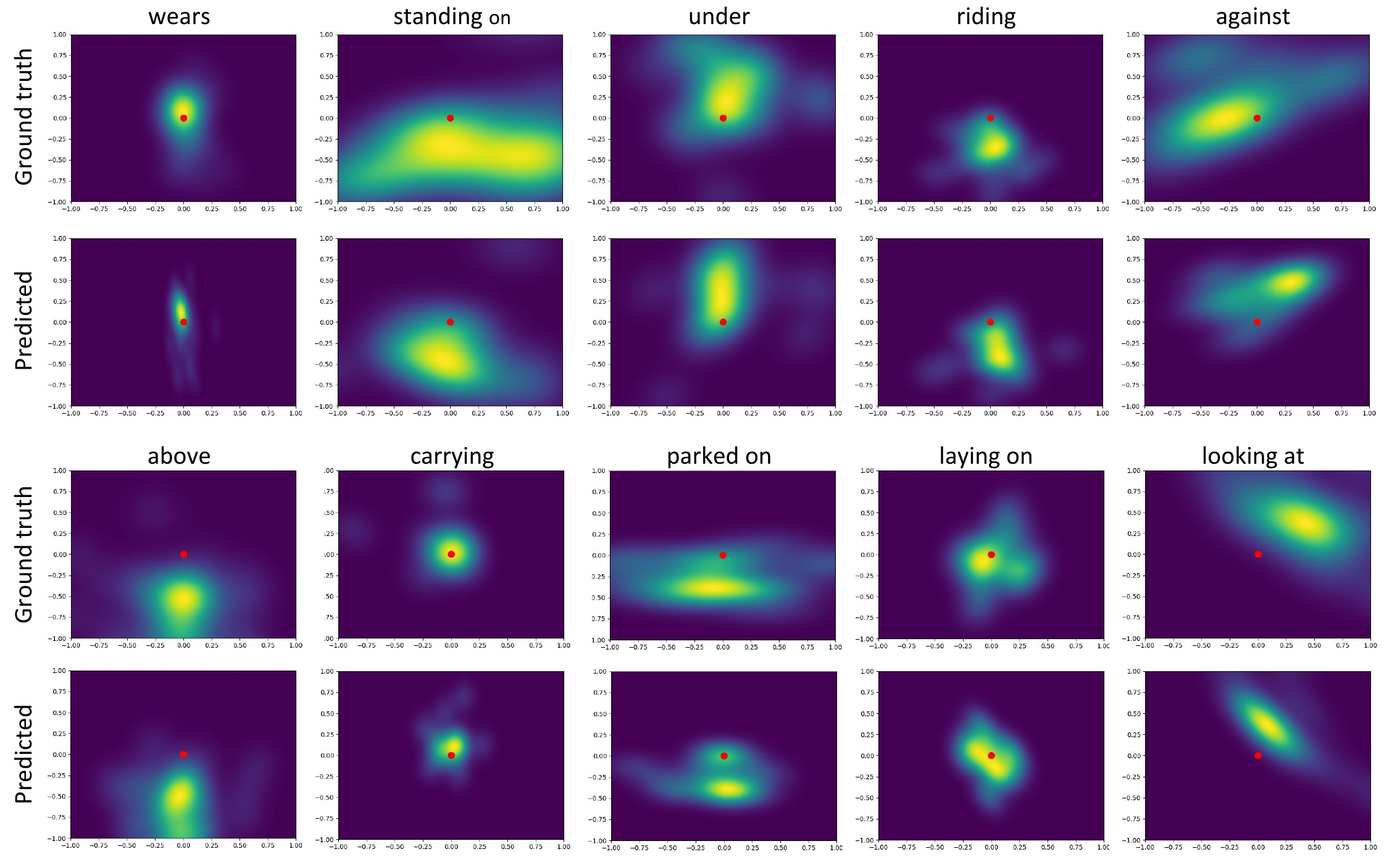}
\caption{\textbf{Heatmaps generated from object and subject relative positions for selected predicate categories.} The object in each image is centered at point $(0,0)$ and the relative position of the subject is calculated. The heatmaps are generated from the relative distances of centers of object and subject. Top: Ground truth boxes. Bottom: our predicted boxes (after masking the location information from the graph representation and letting it be synthesized.} %
\label{fig:heatmaps}
\end{figure*}

\paragraph{Spatial distribution of predicates} Figure \ref{fig:heatmaps} visualizes the heatmaps of the ground truth and predicted bounding box distributions per predicate. For every triplet (\ie subject - predicate - object) in the test set we predict the subject and object bounding box coordinates $\hat{x}_i$. From there, for each triplet we extract the relative distance between the object and subject centers, which are then grouped by predicate category. The plot shows the spatial distribution of each predicate. We observe similar distributions, in particular for the spatially well-constrained relationships, such as \texttt{wears}, \texttt{above}, \texttt{riding}, etc. This indicates that our model has learned to accurately localize new (predicted) objects in relation to objects already existing in the scene.  

\paragraph{User interface video} This supplement also contains a video, demonstrating a user interface for interactive image manipulation. In the video one can see that our method allows multiple changes in a given image. \url{https://he-dhamo.github.io/SIMSG/}

\begin{figure*}[t]
\centering
\includegraphics[width=\linewidth]{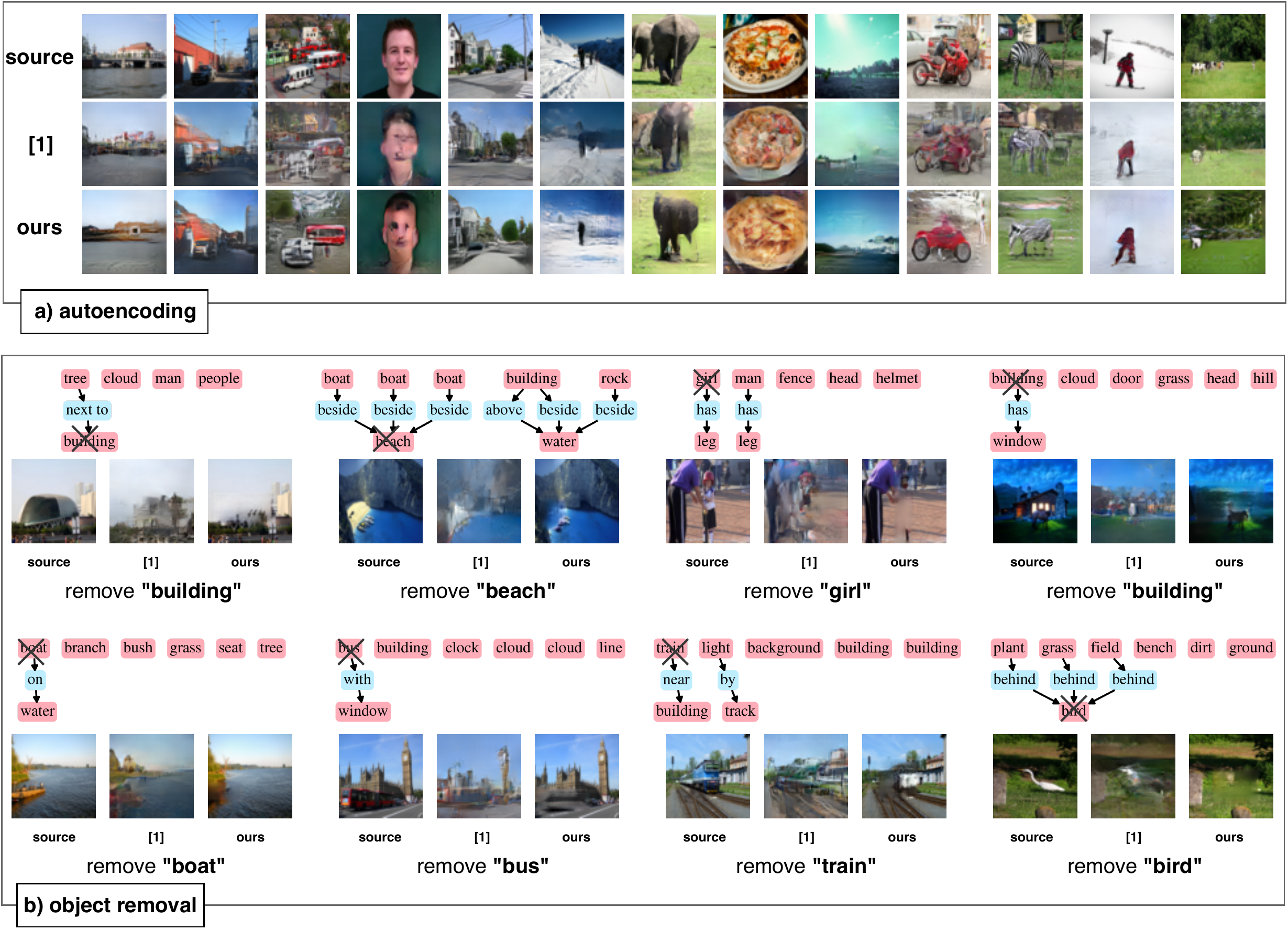}
\caption{\textbf{Qualitative results comparing ours CRN and \cite{ashual2019specifying}} \textbf{a)} Fully-generative setting \textbf{b)} Object removal} 
\label{fig:comparison}
\end{figure*}

\paragraph{Comparison}
Figure \ref{fig:comparison} presents qualitative samples of our method and a comparison to \cite{ashual2019specifying} for the auto-encoding (a) and object removal task (b). 
We adapt \cite{ashual2019specifying} for object removal by removing a node and its connecting edges from the input graph (same as in ours), while the visual features of the remaining nodes (coming from our source image) are used to reconstruct the rest of the image. 
We achieve similar results for the auto-encoding, even though our method is not specifically trained for the fully-generative task. 
As for object removal, our method performs generally better, since it is intended for direct manipulation on an image. 
For a fair comparison, in our experiments, we train \cite{ashual2019specifying} on Visual Genome. Since Visual Genome lacks segmentation masks, we disable the mask discriminator. For this reason, we expect lower quality results than presented in the original paper (trained on MS-COCO with mask supervision and simpler scene graphs).

\subsection{Ablation study on CLEVR}
Tables \ref{tab:clevr_generation_supp} and \ref{tab:clevr_changes_supp} provide additional results on CLEVR, namely for the image reconstruction and manipulation tasks. We observe that the version of our method with a SPADE decoder outperforms the other models in the reconstruction setting. As for the manipulation modes, our method clearly dominates for relationship changes, while the performance for other changes is similar with the baseline.  

\begin{table*}[t]
    \centering
    \footnotesize %
    \begin{tabular}{l@{\hskip 9pt} c@{\hskip 20pt} c@{\hskip 12pt} c@{\hskip 12pt} c@{\hskip 12pt} c@{\hskip 12pt} c@{\hskip 12pt} c@{\hskip 12pt} c}
    \toprule
    \multirow{2}{*}{Method} & \multirow{2}{*}{Decoder} & \multicolumn{5}{c}{All pixels} & \multicolumn{2}{c}{RoI only} \\
    \cmidrule(l{1pt}r{9pt}){3-6} \cmidrule(l{1pt}r{3pt}){7-8}
     & & MAE $\downarrow$ & SSIM $\uparrow$ & LPIPS $\downarrow$ & FID $\downarrow$ & MAE $\downarrow$ & SSIM $\uparrow$ \\
    \midrule
    Image Resolution & \multicolumn{7}{c}{$64 \times 64$}\\
    \midrule
    Fully-supervised & CRN & $6.74$ & $97.07$ & $0.035$ & $5.34$ & $9.34$ & $93.49$\\
    Ours (GT) \quad w/o $\phi_i$ & CRN & $7.96$ & $97.92$ & $0.016$ & $4.52$ & $14.36$ & $81.75$ \\
    Ours (GT) \quad w/ $\phi_i$ & CRN & $6.15$ & $98.50$ & $0.008$ & $3.73$ & $10.47$ & $88.53$\\
    Ours (GT) \quad w/o $\phi_i$ & SPADE & $4.25$ & $98.79$ & $0.009$ & $3.75$ & $9.67$ & $87.13$ \\
    Ours (GT) \quad w/ $\phi_i$ & SPADE &  $2.73$ & $99.35$ & $0.002$ & $3.42$ & $5.42$ & $94.16$\\
    \midrule
    Image Resolution & \multicolumn{7}{c}{$128 \times 128$}\\
    \midrule
    Fully-supervised & CRN & $9.83$ & $97.36$ & $0.061$ & $4.42$ & $12.38$ & $91.94$\\
    Ours (GT) \quad w/o $\phi_i$ & CRN &  $14.82$ & $96.85$ & $0.041$ & $8.09$ & $20.59$ & $74.71$ \\
    Ours (GT) \quad w/ $\phi_i$ & CRN & $14.47$ & $96.93$ & $0.038$ & $8.36$ & $19.56$ & $75.25$\\
    Ours (GT) \quad w/o $\phi_i$ & SPADE & $9.26$ & $98.27$ & $0.029$ & $3.21$ & $15.74$ & $79.81$ \\
    Ours (GT) \quad w/ $\phi_i$ & SPADE &  $5.39$ & $99.18$ & $0.007$ & $1.17$ & $8.32$ & $89.84$\\
    \bottomrule
\end{tabular}%
\vspace{5pt}
    \caption{\textbf{Image reconstruction on CLEVR.} We report the results using ground truth scene graphs (GT).}
    \label{tab:clevr_generation_supp}
\end{table*}

\begin{table*}[t]
    \centering
    \footnotesize %
    \begin{tabular}{l@{\hskip 9pt} c@{\hskip 20pt} c@{\hskip 12pt} c@{\hskip 12pt} c@{\hskip 12pt} c@{\hskip 12pt} c@{\hskip 12pt} c}
    \toprule
    \multirow{2}{*}{Method} & \multirow{2}{*}{Decoder} & \multicolumn{4}{c}{All pixels} & \multicolumn{2}{c}{RoI only} \\
    \cmidrule(l{1pt}r{9pt}){3-5} \cmidrule(l{1pt}r{3pt}){6-7}
     & & MAE $\downarrow$ & SSIM $\uparrow$ & LPIPS $\downarrow$ & MAE $\downarrow$ & SSIM $\uparrow$ \\
    \midrule
    \midrule
    Image Resolution & \multicolumn{6}{c}{$64 \times 64$}\\
    \midrule
    \midrule
    Change Mode & \multicolumn{6}{c}{Addition}\\
    \midrule
    Fully-supervised & CRN & $6.57$ & $98.60$ & $0.013$ & $7.68$ & $97.72$\\
    Ours (GT) \quad w/ $\phi_i$ & CRN & $7.88$ & $96.93$ & $0.027$ & $9.79$ & $95.10$\\
    Ours (GT) \quad w/ $\phi_i$ & SPADE &  $4.96$ & $97.45$ & $0.026$ & $6.13$ & $96.86$\\
    \midrule
    Change Mode & \multicolumn{6}{c}{Removal}\\
    \midrule
    Fully-supervised & CRN & $4.52$ & $98.60$ & $0.006$ & $5.53$ & $97.17$\\
    Ours (GT) \quad w/ $\phi_i$ & CRN & $5.67$ & $97.13$ & $0.026$ & $7.02$ & $96.41$\\
    Ours (GT) \quad w/ $\phi_i$ & SPADE &  $3.45$ & $97.32$ & $0.022$ & $3.88$ & $98.09$\\
    \midrule
    Change Mode & \multicolumn{6}{c}{Replacement}\\
    \midrule
    Fully-supervised & CRN & $6.64$ & $97.76$ & $0.015$ & $7.33$ & $97.11$\\
    Ours (GT) \quad w/ $\phi_i$ & CRN & $8.24$ & $96.96$ & $0.025$ & $9.29$ & $96.02$\\
    Ours (GT) \quad w/ $\phi_i$ & SPADE &  $5.88$ & $97.43$ & $0.023$ & $6.56$ & $97.48$ \\
    \midrule
    Change Mode & \multicolumn{6}{c}{Relationship changing}\\
    \midrule
    Fully-supervised & CRN & $9.76$ & $93.91$ & $0.111$ & $17.51$ & $83.24$\\
    Ours (GT) \quad w/ $\phi_i$ & CRN & $10.09$ & $93.50$ & $0.0678$ &  $14.91$ & $86.17$\\
    Ours (GT) \quad w/ $\phi_i$ & SPADE &  $8.11$ & $93.75$ & $0.069$ & $13.01$ & $86.99$\\
    \midrule
    \midrule
    Image Resolution & \multicolumn{6}{c}{$128 \times 128$}\\
    \midrule
    \midrule
    Change Mode & \multicolumn{6}{c}{Addition}\\
    \midrule
    Fully-supervised & CRN & $9.72$ & $97.57$ & $0.031$ & $10.61$ & $94.09$\\
    Ours (GT) \quad w/ $\phi_i$ & CRN & $13.77$ & $96.44$ & $0.048$ & $13.21$ & $91.05$\\
    Ours (GT) \quad w/ $\phi_i$ & SPADE &  $7.79$ & $97.89$ & $0.040$ & $7.57$ & $96.18$\\
    \midrule
    Change Mode & \multicolumn{6}{c}{Removal}\\
    \midrule
    Fully-supervised & CRN & $6.15$ & $98.72$ & $0.014$ & $7.27$ & $95.58$\\
    Ours (GT) \quad w/ $\phi_i$ & CRN & $11.75$ & $97.21$ & $0.052$ &  $11.55$ & $92.34$\\
    Ours (GT) \quad w/ $\phi_i$ & SPADE &  $4.48$ & $98.54$ & $0.042$ & $4.60$ & $97.68$\\
    \midrule
    Change Mode & \multicolumn{6}{c}{Replacement}\\
    \midrule
    Fully-supervised & CRN & $10.49$ & $97.57$ & $0.035$ & $11.23$ & $95.09$\\
    Ours (GT) \quad w/ $\phi_i$ & CRN & $16.38$ & $96.14$ & $0.052$ & $14.74$ & $91.98$\\
    Ours (GT) \quad w/ $\phi_i$ & SPADE &  $10.25$ & $97.51$ & $0.041$ &  $9.98$ & $96.14$\\
    \midrule
    Change Mode & \multicolumn{6}{c}{Relationship changing}\\
    \midrule
    Fully-supervised & CRN & $13.91$ & $95.26$ & $0.169$ & $21.49$ & $82.46$\\
    Ours (GT) \quad w/ $\phi_i$ & CRN & $16.61$ & $94.60$ & $0.128$ & $19.21$ & $85.24$\\
    Ours (GT) \quad w/ $\phi_i$ & SPADE &  $11.62$ & $95.76$ & $0.125$ & $14.01$ & $89.15$\\
    \bottomrule
\end{tabular}%
\vspace{5pt}
    \caption{\textbf{Image manipulation on CLEVR.} We report the results for different categories of modifications.}
    \label{tab:clevr_changes_supp}
\end{table*}

\subsection{Datasets}
\paragraph{CLEVR~\cite{johnson2017clevr}.} We generate 21,310 pairs of images which we split into $80\%$ for training, $10\%$ for validation and $10\%$ for testing. Each data pair illustrates the same scene under a specific change, such as position swapping, addition, removal or changing the attributes of the objects. The images are of size $128\times128\times3$ and contain $n$ random objects ($3 \leq n \leq 7$) with random shapes and colors. Since there are no graph annotations, we define predicates as the relative positions \{\texttt{in front of}, \texttt{behind}, \texttt{left of}, \texttt{right of}\} of different pairs of objects in the scene. The generated dataset includes annotated information of scene graphs, bounding boxes, object classes and object attributes. 

\paragraph{Visual Genome (VG)~\cite{krishna2017visual}.} We use the VG v1.4 dataset with the splits as proposed in \cite{johnson2018image}. The training, validation and test set contain namely 80\%, 10\% and 10\% of the dataset. After applying the pre-processing of \cite{johnson2018image} the dataset contains 178 object categories and 45 relationship types. The final dataset after processing comprises 62,565 train, 5,506 val, and 5,088 test images with graphs annotations. We evaluate our models with GT scene graphs on all the images of the test set. For the experiments with predicted scene graphs (P), an image filtering takes place (\eg no objects are detected), therefore the evaluation in performed in 3874 images from the test set. We observed relationship duplicates in the dataset and we empirically found that it does not affect the image generation task. However, it leads to ambiguity on modification time (when tested with GT graphs) once we change only one of the duplicate edges. Therefore, we remove such duplicates once one of them is edited. 

\subsection{Implementation details}

\subsubsection{Image $\rightarrow$ scene graph}

A state-of-the-art scene graph prediction network~\cite{li2018factorizable} is used to acquire scene graphs for the experiments on VG. We use their publicly available implementation\footnote{\url{https://github.com/yikang-li/FactorizableNet}} to train the model. 
The data used to train the network is pre-processed following~\cite{dai2017detecting}, resulting in a typically used subset of Visual Genome (sVG) that includes 399 object and 24 predicate categories. We then split the data as in~\cite{johnson2018image} to avoid overlap in the training data for the image manipulation model. We train the model for 30 epochs with a batch size of 8 images using the default settings from~\cite{li2018factorizable}. 

\subsubsection{Scene graph $\rightarrow$ image}

\paragraph{SGN architecture details.} The learned embeddings of the object $c_i$ and predicate $r_i$ both have $128$ dimensions. We create the full representation of each object $o_i$ by concatenating $c_i$ together with the bounding box coordinates $x_i$ (top, left, bottom, right) and the visual features (n=128) corresponding to the cropped image region defined by the bounding box. 
The features are extracted by a VGG-16 architecture~\cite{simonyan2014very} followed by a 128-dimensional fully connected layer. 

During training, to hide information from the network, we randomly mask the visual features $\phi_i$ and/or object coordinates $x_i$ with independent probabilities of $p_{\phi}=0.25$ and $p_x=0.35$. 

The SGN consists of $5$ layers. $\tau_e$ and $\tau_n$ are implemented as $2$-layer MLPs with 512 hidden and 128 output units. %
The last layer of the SGN returns the outputs; the node features (s=128), binary masks ($16\times16$) and bounding box coordinates by $2$-layer MLP with a hidden size of $128$ (which is needed to add or re-position objects).

\paragraph{CRN architecture details.} The CRN architecture consists of $5$ cascaded refinement modules, with the output number of channels being 1024, 512, 256, 128 and 64 respectively. 
Each module consists of two convolutions ($3\times3$), each followed by batch normalization~\cite{ioffe2015batch} and leaky Relu. The output of each module is concatenated with a down-sampled version of the initial input to the CRN. The initial input is the concatenation of the predicted layout and the masked image features. The generated images have a resolution of $64\times64$.

\paragraph{SPADE architecture details.} The SPADE architecture used in this work contains $5$ residual blocks. The output number of channels is namely 1024, 512, 256, 128 and 64. In each block, the layout is fed in the SPADE normalization layer, to modulate the layer activations, while the image counterpart is concatenated with the result. The global discriminator $D_{global}$ contains two scales. 

The object discriminator in both cases is only applied on the image areas that have changed, \ie have been in-painted.

\paragraph{Full-image branch details.} The image regions that we randomly mask during training are replaced by Gaussian noise. Image features are extracted using 32 convolutional filters ($1\times1$), followed by batch normalization and Relu activation. Additionally, a mask is concatenated with the image features that is $1$ in the regions of interest (noise) and $0$ otherwise, so that the areas to be modified are easier for the network to identify.

\paragraph{Training settings.} In all experiments presented in this paper, the models were trained with Adam optimization~\cite{kingma2014adam} with a base learning rate of $10^{-4}$. The weighting values for different loss terms in our method are shown in Table~\ref{tab:lambdas}. The batch size for the images in $64 \times 64$ resolution is $32$, while for $128 \times 128$ is 8. All objects in an image batch are fed at the same time in the object-level units, \ie SGN, visual feature extractor and discriminator. 

All models on VG were trained for 300k iterations and on CLEVR for 40k iterations. Training on an Nvidia RTX GPU, for images of size $64 \times 64$ takes about 3 days for Visual Genome and 4 hours for CLEVR.

\begin{table}[h]
    \centering
    \begin{tabular}{l|cc}
    \toprule
        Loss factor & Weight CRN & Weight SPADE \\
        \midrule
       $\lambda_g$  & $0.01$ & $1$ \\\ %
       $\lambda_o$  & $0.01$ & $0.1$ \\ %
       $\lambda_a$  & $0.1$ & $0.1$\\ %
       $\lambda_b$ & 10 & 50 \\
        $\lambda_f$ & - & 10 \\
        $\lambda_p$ & - & 10 \\
       \bottomrule
    \end{tabular}
    \vspace{2mm}
    \caption{Loss weighting values}
    \label{tab:lambdas}
\end{table}

\subsection{Failure cases}

\begin{figure}[!h]
\centering
\includegraphics[width=0.85\linewidth]{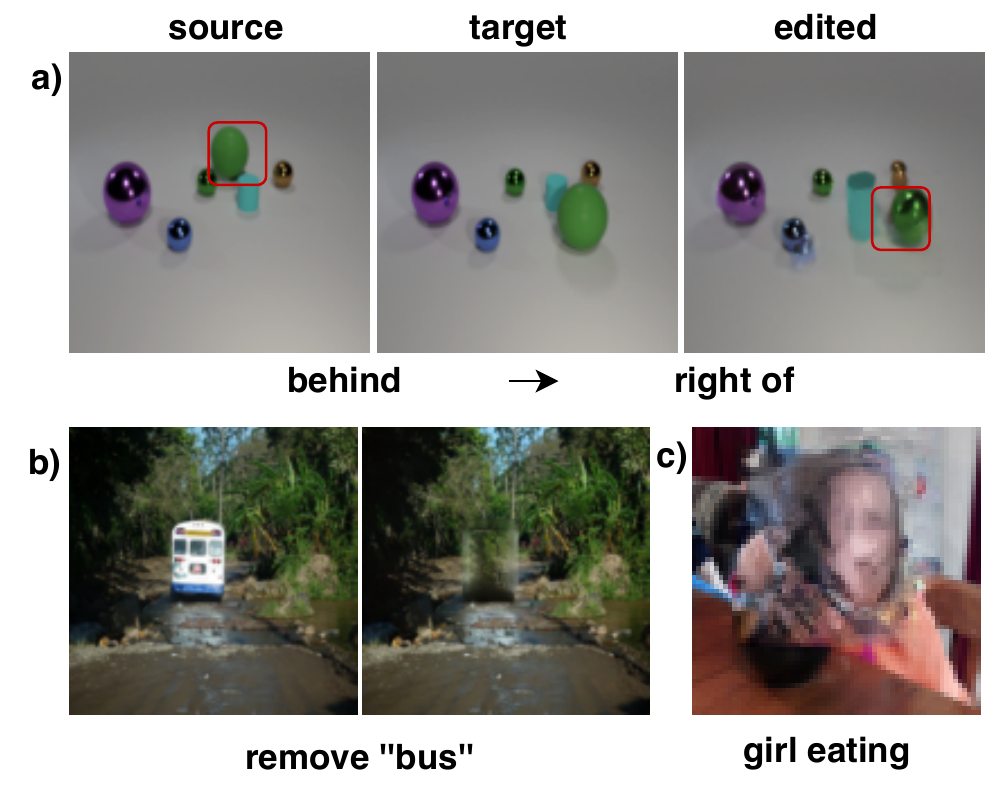}
\caption{Illustration of failure cases.} 
\label{fig:failure}
\end{figure}

In the proposed image manipulation task we have to restrict the feature encoding to prevent the encoder from ``copying'' the whole RoI, which is not desired if, for instance, we want to re-position non-rigid objects, e.g.~from \texttt{sitting} to \texttt{standing}. While the model is able to retain general appearance information such as colors and textures, it is true that, as a side effect some visual properties of modified objects are not recovered. For instance, the color of the green object in Figure \ref{fig:failure} a) is preserved but not the material. 

The model does not adapt unchanged areas of the image as a consequence of a change in the modified parts. For example, shadows or reflections do not follow the re-positioned objects, if those are not nodes of the graph and explicitly marked as changing subject by the user, Figure \ref{fig:failure} b).

In addition, similarly to other methods evaluated on Visual Genome, the quality of some close objects remains limited, \eg close-up of people eating, Figure \ref{fig:failure} c).
Also, having a node \texttt{face} on animals, typically gives them a human face.

